\documentclass[conference]{IEEEtran}

\usepackage{cite}

\ifCLASSINFOpdf

\else
\fi

\usepackage{amsfonts}
\usepackage{graphicx}
\usepackage[utf8]{inputenc} %
\usepackage[T1]{fontenc}    %
\usepackage{hyperref}       %
\usepackage{url}            %
\usepackage{booktabs}       %
\usepackage{amsfonts}       %
\usepackage{nicefrac}       %
\usepackage{microtype}      %
\usepackage{xcolor}         %
\usepackage{algorithm}
\usepackage{algorithmic}
\usepackage{amssymb}
\usepackage{amsmath}
\usepackage{subfig}

\newenvironment{packed_itemize}{
\begin{itemize}
  \setlength{\itemsep}{1pt}
  \setlength{\parskip}{0pt}
  \setlength{\parsep}{0pt}
}{\end{itemize}}

\begin{document}
\title{Learning semantic image quality for fetal ultrasound from noisy ranking annotation}

\author{\IEEEauthorblockN{Manxi Lin$^{1,*}$, Jakob Ambsdorf$^{2,4,*}$, Emilie Pi Fogtmann Sejer$^{3}$, Zahra Bashir$^{3}$, Chun Kit Wong$^{1,4}$,\\ Paraskevas Pegios$^{1,4}$, Alberto Raheli$^{1}$, Morten Bo Søndergaard Svendsen$^{2,3}$, Mads Nielsen$^{2,4}$,\\ Martin Grønnebæk Tolsgaard,$^{3}$, Anders Nymark Christensen$^{1}$, Aasa Feragen$^{1,4}$}
\IEEEauthorblockA{$^{1}$DTU Compute, Technical University of Denmark, Kgs Lyngby, Denmark\\
$^{2}$Department of Computer Science, University of Copenhagen, Copenhagen, Denmark\\
$^{3}$Copenhagen Academy of Medical Education and Simulation (CAMES), Rigshospitalet, Copenhagen, Denmark\\
$^{4}$Pioneer Centre for AI, Copenhagen, Denmark\\
Email: \{manli, afhar\}@dtu.dk, jaam@di.ku.dk; $^{*}$Equal contribution}
}

\maketitle

\thispagestyle{plain}
\pagestyle{plain}

\begin{abstract}
We introduce the notion of semantic image quality for applications where image quality relies on semantic requirements. Working in fetal ultrasound, where ranking is challenging and annotations are noisy, we design a robust coarse-to-fine model that ranks images based on their semantic image quality and endow our predicted rankings with an uncertainty estimate. To annotate rankings on training data, we design an efficient ranking annotation scheme based on the merge sort algorithm. Finally, we compare our ranking algorithm to a number of state-of-the-art ranking algorithms on a challenging fetal ultrasound quality assessment task, showing the superior performance of our method on the majority of rank correlation metrics. 
\end{abstract}

\IEEEpeerreviewmaketitle

\section{Introduction}

Image quality is crucial for an array of medical imaging tasks~\cite{ozer2023explainable, lyu2021ultrasound}, but acquiring high-quality images is challenging in terms of cost, time, equipment, and operator skill. While in computer vision, image quality tends to encapsulate low-level properties like sharpness or artefacts~\cite{chen2022fine}, the quality of medical images like fetal ultrasound pose complex requirements on semantic image content~\cite{lyu2021ultrasound, zarb2010image}. Here, images of "sufficient quality"  satisfy a set of standardized requirements~\cite {salomon2019isuog}, leading us to define a new \emph{semantic image quality} paradigm.

We consider image quality assessment as a ranking task, as opposed to existing classification of images as either good enough or not, via regression on a global score. Classification is too coarse, as the "good enough" image might be unattainable if the patient is too ill to sit still; if the operator is not sufficiently skilled, or if high patient BMI or fetal position inhibits high-quality acquisition. We still want the best image quality possible, which requires finely ranking one image as better than another. Moreover, applications such as navigation and auto-capture for ultrasound acquisition assistance also depend on detecting small improvements in image quality. 

\begin{figure}[t]
  \centering
  \includegraphics[width=0.9\linewidth]{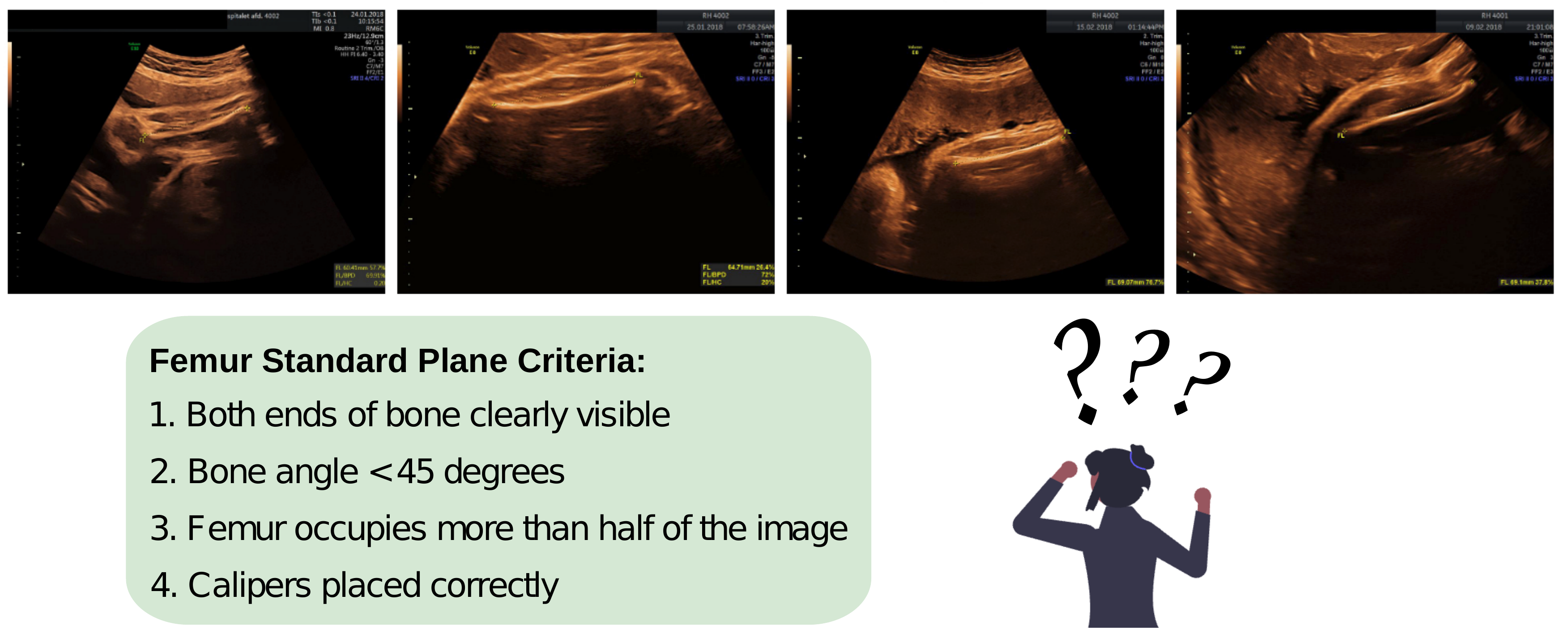}
  \caption{Fetal ultrasound \emph{standard planes} are particular 2D cross sections of the fetal anatomy, defined by the presence, absence, and visual quality of specific organs. Above, 4 femur planes are sorted by their standard plane quality -- which is challenging, as the assessment depends on multiple simultaneous criteria.}
  \label{fig:teaser}
\end{figure}
While this could be done by regressing on a quality score, we hypothesize that annotated scores or categories can be imprecise and biased~\cite{gao2015learning}, as the complexity of semantic image quality makes the task naturally challenging. We observe this through intra-observer inconsistency in annotated scores and rankings, also supported by previous work~\cite{zarb2010image}. However, when instead annotating rankings, the annotators do not need to sense the scale of sample difference during annotation. The annotators are thus asked to perform a simpler task - instead of 'how much this image is better than the other', the annotators only need to answer 'is this image better than the other'. This is supported by literature on human decision-making, which indicates that humans make better assessments when asked to rank than to score. Motivated by this, we collected image quality rankings rather than image quality scores~\cite{harzing2009rating}, and learned image \emph{ranking} directly from \emph{ranking} annotations, rather than rank images implicitly by training regression or classification models.

 \textbf{In this paper, we studied the problem of ranking images according to their semantic quality. We contribute:}
 \begin{packed_itemize}
    \item Instead of classifying images as "good enough" or not, we propose an \textbf{ordinal-regression-based network} to learn semantic image quality ranking in a coarse-to-fine manner.
     \item Modelling of the predictions' \textbf{ranking uncertainty}.
     \item Validation against state-of-the-art on a fetal ultrasound image quality assessment task.
     \item In order to train algorithms for ranking, we propose a \textbf{ranking annotation scheme} based on the merge sort algorithm, enabling easy annotation of ranking.  
 \end{packed_itemize}

Note that our developed methodology applies to any semantic image quality assessment application.

\section{Related Work}

\emph{Learning-to-rank} (LTR) models sort samples according to their relevance for a given query~\cite{swezey2021pirank}. 
Pointwise~\cite{crammer2001pranking} approaches, which score single samples, suffer from lack of context, whereas pairwise~\cite{burges2005learning} or listwise~\cite{cao2007learning, lan2014position, swezey2021pirank} approaches predict the relative order of pairs or complete lists, respectively.
Validation metrics for listwise ranking, e.g.~Normalized Discounted Cumulative Gain (NDCG), are not differentiable and cannot be used as loss functions. Existing work thus optimize an upper bound~\cite{valizadegan2009learning, yue2007support} or approximate the metric~\cite{swezey2021pirank}. SoDeep uses a sorting approximation of non-differentiable metrics trained on synthetic data~\cite{engilberge2019sodeep}. %
While LTR models are popular for image quality assessment~\cite{gao2015learning, ma2017dipiq, liu2017rankiqa}, they are mostly used as auxiliary tasks for regressing quality scores. An exception is~\cite{gao2015learning}, which generates different levels of image corruption and learns image quality from pairwise ranking.
For ultrasound, LTR has been used as an auxiliary loss for regression~\cite{chen2022fine,lyu2021ultrasound}, but synthetic noise is difficult to generate for fetal ultrasound quality assessment~\cite{lyu2021ultrasound}. We instead learn to rank semantic image quality directly from noisy ranking annotations.  

\emph{Relative attribute learning} quantifies the strength of attributes relative to other images~\cite{parikh2011relative}. This is exemplified by attributes such as 'young', 'natural', and 'smiling', which are easy to compare between two images, but harder to numerically score. Existing approaches include siamese networks optimized with a Hinge or rank SVM loss~\cite{yang2016deep,ahmed2021relative}; pairwise ranking treated as binary classification as in RankNet~\cite{souri2017deep}; or learning robust relative attributes from noisy data by excluding noisy pairs~\cite{xu2020not}. Our \emph{semantic image quality} is a relative attribute. Relative attributes are usually modeled for sampled subsets, often pairs. We, conversely, learn to rank semantic image quality from full ranking annotations of the dataset, coupled with an approach to assess the uncertainty of the learned rankings.  

\begin{figure}[b]
  \centering
  \includegraphics[width=\linewidth]{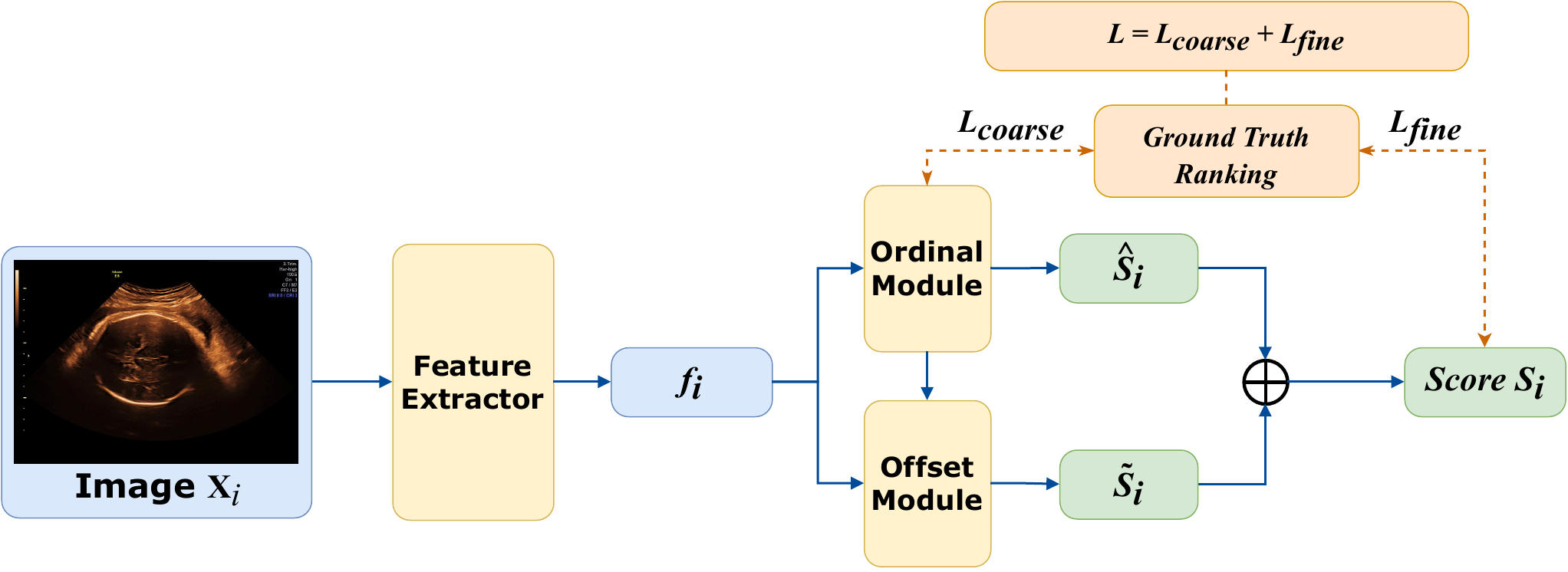}
  \caption{Network architecture of the proposed ORBNet.}
  \label{fig:orbnet}
\end{figure}

\emph{Ordinal regression}, situated between regression and classification, predicts pointwise ordered class labels, rather than determining the ranking of a set. Classical approaches reformulate ordinal regression as multiple binary classification tasks~\cite{NIPS2001_5531a583,shen2005ranking,  chu2005new, cao2012human, antoniuk2016v}, but 
these methods cannot guarantee ordinal consistency. 
In~\cite{cao2020rank} weight-sharing constraints are imposed in the output layer of a deep network, giving theoretical guarantees for consistent confidence scores and regularization. In~\cite{garg2020robust} a robust deep ordinal regression method is proposed along with loss correction for class-dependent label noise~\cite{patrini2017making}. We use ordinal regression to learn a coarse global ranking.

\section{Method}

\subsection{Learning to Rank with ORBNet}
\label{sec:orbnet}

We aim to rank a dataset $D = \{ \mathbf{x}_i, y_i \}_{i=i}^{n}$, where $\mathbf{x}_i$ is the $i$-th image and $y_i \in \mathbb{N}$ its annotated rank. For simplicity, we assume no ties, i.e. $y_i \neq y_j \ \forall \ i \neq j$. We tackle this by learning an intermediate score $\mathbf{x}_i \mapsto s_i$, which is easily converted to a rank $\hat{y}_i$.
For this, we propose an ordinal-regression-based neural network (ORBNet) (Fig~\ref{fig:orbnet}), with three parts: (1) a pre-trained feature extractor $\theta$ gathering high-level semantic features $f_i$ from $\mathbf{x}_i$ (see Sec.~\ref{sec:data}), (2) an ordinal module $\psi$ to get a coarse, global ranking of the extracted features, and (3) an offset module $\zeta$ giving a fine-grained, local scoring.

\paragraph*{Coarse global ranking with ordinal regression} We first learn a reliable rough ranking via ordinal regression, by splitting ranking labels into $m$ bins. Each ranking label $y_i$ thus is encoded as a set of binary labels ${\{b_{ij}\}}_{j=1}^{m-1}$, where
\begin{equation}
    b_{ij} = \begin{cases}
    1 & \textrm{if } y_i >=  j\tau \\
    0 & \textrm{otherwise,} \\
    \end{cases}
\end{equation}
where $\tau=\frac{n}{m}$ is the range of each bin. The ordinal mapping $\psi \colon f_i \mapsto l_i$ maps to a coarse global ranking $l_i \in \mathbb{R}^{m-1}$, binning the image $x_i$ within the whole dataset. Each coordinate of $l_i$, if activated by the sigmoid $\sigma$, represents the probability of $x_i$ larger than the left bound of the bin. We train $l_i$ to align with ${\{b_{ij}\}}_{j=1}^{m-1}$ using binary cross-entropy $l_{coarse}^i = -\sum_{j=1}^{m-1}(log(\sigma(l_{ij}))\cdot b_{ij} + (1-log(\sigma(l_{ij})))\cdot (1 - b_{ij}))$,
where $l_{ij}$ is the $j$-th element of $l_i$. The bin to which the image is assigned defines a score
$\bar{s}_i = \tau \cdot \sum_{j=1}^{m-1}l_{ij}$,
which is within $[0, \frac{n(m-1)}{m}]$, i.e., the left bound of the bin the image belongs to, in a normalized scale. We implement $\psi$ as a multi-layer perceptron (MLP); the output layer biases are not shared~\cite{cao2020rank}.

\paragraph*{Fine-tune local ranking by learning an offset} The effect of noise is reduced through our rough binned ranking, but we also lose fine-grained information. We therefore add an offset module $\zeta$ learning small bin-wise offsets. The offset module $\zeta \colon \{f_i, l_i\} \mapsto \widetilde{s}$ is an MLP taking the concatenation $(f_i,l_i)$ as input, outputting an offset $\widetilde{s} \in \mathbb{R}$ which is limited to a local range, to retain the effect of the predicted global order. This gives a final predicted quality score $s_i = \bar{s}_i + \tau\cdot\sigma(\widetilde{s}) \in [0, 1]$- for $x_i$, which is used to train the offset layer with RankNet~\cite{burges2005learning}. For each pair of images in $D$, $l_{fine}^{(i,j)} = -(log(\sigma(s_i - s_j))p_{ij} + (1-log(\sigma(s_i - s_j)))(1-p_{i,j}))$,
where
\begin{equation}
    p_{ij} = \begin{cases}
        1, & y_i > y_j\\ 
        0, & y_i < y_j\\
        0.5, & y_i = y_j
    \end{cases}. 
\end{equation}
The ORBNet is trained end-to-end using $l = l_{coarse} + l_{fine}$.

\subsection{Assessing Ranking Uncertainty}
\label{sec:ensemble}

As our ranking annotations are noisy, we also expect uncertain predictions. While the variance of the predicted ranks is easy to quantify, this would not help the user, as the position of an image in the quality-ranked dataset is not clinically relevant. Instead, we choose to assess the uncertainty of binary decisions of the form "which of two images is better?"

We evaluate the uncertainty of pairwise ORBNet rankings using %
MC Dropout~\cite{gal2016dropout} at inference time. Given a pair of images $\mathbf{x}^a$ and $\mathbf{x}^b$, we use $N$ forward passes of the model with Dropout enabled, to obtain multiple %
predicted ranking scores $\widetilde{y}_{a}$ for each image $\mathbf{x}^a$. The share of predicted rankings $\mathbf{x}^a$ over $\mathbf{x}^b$ in the form of a Bernoulli distribution $P(\widetilde{y}_{a} > y_b)$ is then interpreted as a confidence in ranking $\mathbf{x}_a$ over $\mathbf{x}_b$.

\subsection{Annotating Rankings of Entire Datasets via Merge Sort}
\label{subsec:annotation_scheme}
Considering the combinatorial complexity that is introduced when ranking entire lists, annotating complete dataset rankings is expensive, as every pair of items needs to be compared. In this paper, we propose a scheme for annotating a full dataset ranking based on the merge sort algorithm, illustrated in Fig.~\ref{fig:mergesort}:

\begin{figure}
  \centering
  \includegraphics[width=0.7\linewidth]{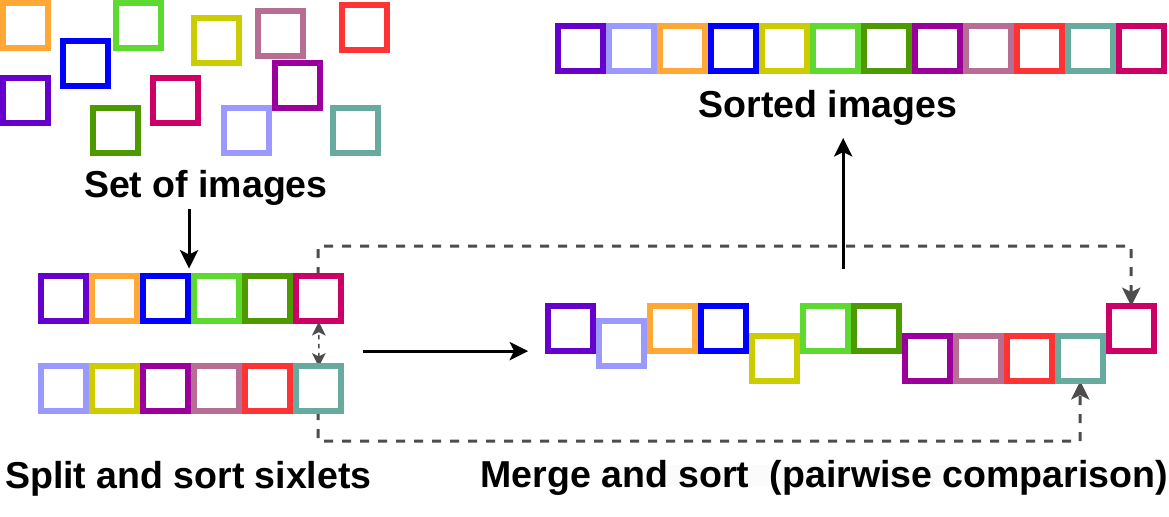}
  \caption{Illustration of the annotation process.}
  \label{fig:mergesort}
\end{figure}

\begin{enumerate}
    \item Divide the original dataset into sub-sets of size $n$, where $n$ is a controllable hyperparameter;
    \item Sort each sub-set;
    \item Until the entire dataset is merged, do:
    \begin{enumerate}
        \item Select the two highest ranking instances from two adjacent sub-sets;
        \item Pick the higher ranking instance of the two, remove it from the sub-set and append it to a new list;
        \item When all sub-sets have been merged, continue by merging the new larger lists;
    \end{enumerate}
\end{enumerate}

The proposed annotation scheme reduces the annotation workload compared to directly sorting an entire list (e.g. in a drag-and-drop interface) by choosing an appropriate value for $n$ and limiting the merge process to an iterative selection of the higher-quality image. The value of $n$ is selected based on (a) fitting all images in sufficient size and resolution on a screen, and (b) not overwhelming the working memory of the annotator (typically $7\pm2$ items~\cite{miller1956magical}), to ensure that the initial ranking can be performed quickly and with a low error-rate. In practice, we determined $n = 6$ to be suitable for our use case on a 27-inch screen. The initial sorting of $n$-lets increases the efficiency of the annotation process by reducing the number of merge iterations compared to starting with image pairs.%

\section{Experiments}
\label{sec:experiments}

\subsection{Data and Implementation} \label{sec:data}
We consider 3rd-trimester fetal ultrasound screening, including 300 femur images, 300 abdomen images, and 264 head images from a national fetal ultrasound screening database~\cite{lin2022saw}. These images were annotated following Sec.~\ref{subsec:annotation_scheme} (annotator A1, MD, Ph.D. fellow in Fetal Medicine: femur and head images, annotator A2, Fetal Medicine consultant, Professor: abdomen images), defining semantic image quality via compliance with the ISUOG guidelines~\cite{salomon2019isuog}. To analyze intra- and inter-annotator ranking consistency, a random sample of 60 head images was additionally annotated by A1 (several days after performing the first trial) as well as by A2.  We kept these 60 head images as a clean test set and trained the model on the rest of the head images, where $10\%$ images were left for validation.

The property bottleneck of a pre-trained progressive concept bottleneck network (PCBM)~\cite{lin2022saw}, which are both representative and human-interpretable via the ISUOG criteria~\cite{salomon2019isuog}, was used as a feature extractor. This generates features ${\{f_i\}}_{i=1}^n$, used as input for the following stages in our ORBNet: The ordinal module is a three-layer MLP including 512, 128, and $m-1=9$ neurons in each hidden layer, with $m=10$ selected empirically. The offset module is another three-layer MLP containing 512, 128, and 1 neurons per hidden layer. The first two layers of the offset module share weights with the ordinal module.       

The ordinal and offset modules were trained with SGD for 30 epochs with momentum 0.9, batch size 32, and L2 regularization 1e-4. The learning rate was initialized with 1e-2 and multiplied by 0.1 after 10 epochs without improvement.
All hyperparameters were selected via ablation experiments.

\begin{figure*}[h!]
	\centering

	\subfloat{\includegraphics[width = 0.16\linewidth, height=0.15\linewidth]{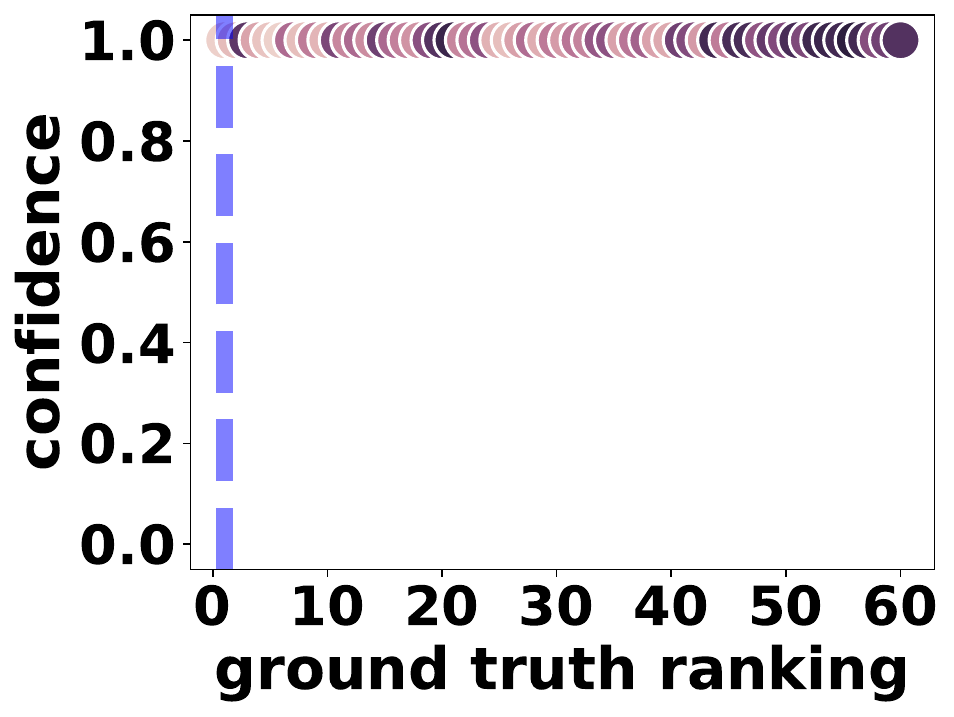}}
	\subfloat{\includegraphics[width = 0.14\linewidth, height=0.15\linewidth]{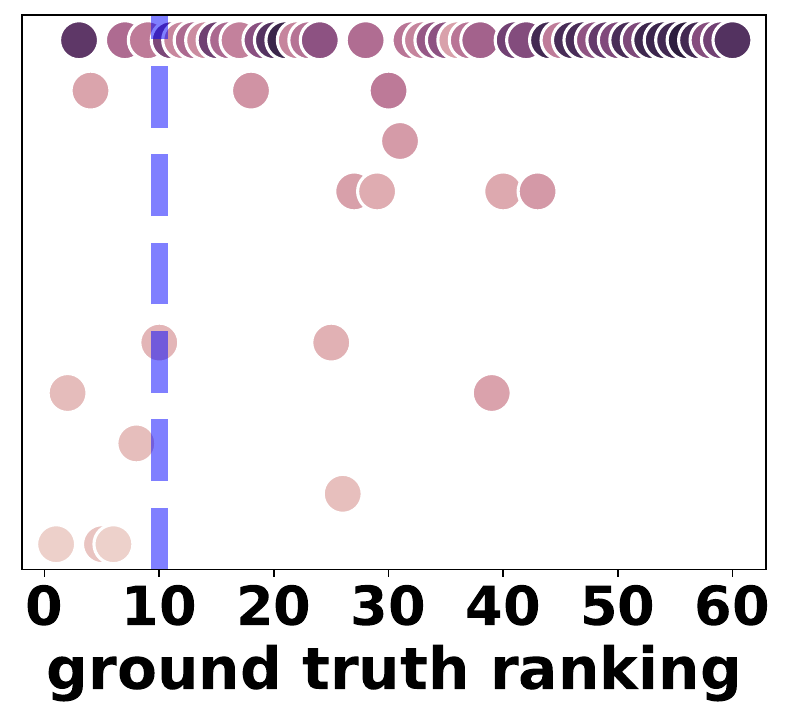}} 
	\subfloat{\includegraphics[width = 0.14\linewidth, height=0.15\linewidth]{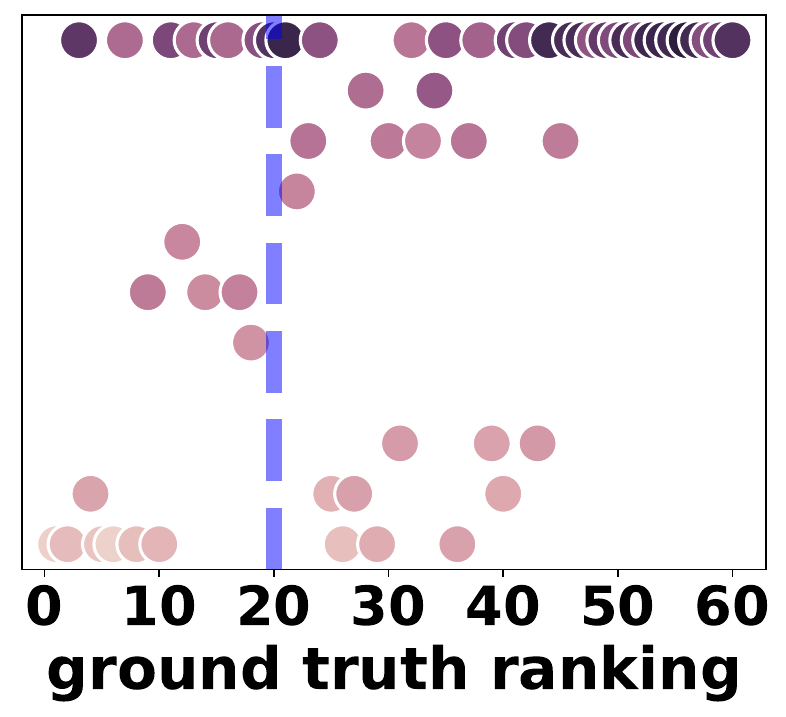}}
	\subfloat{\includegraphics[width = 0.14\linewidth, height=0.15\linewidth]{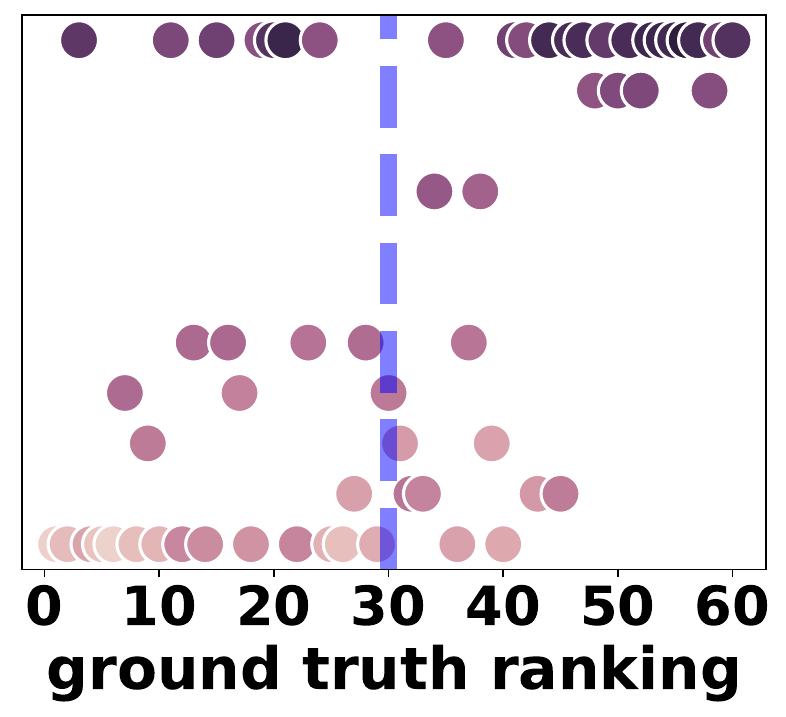}}
	\subfloat{\includegraphics[width = 0.14\linewidth, height=0.15\linewidth]{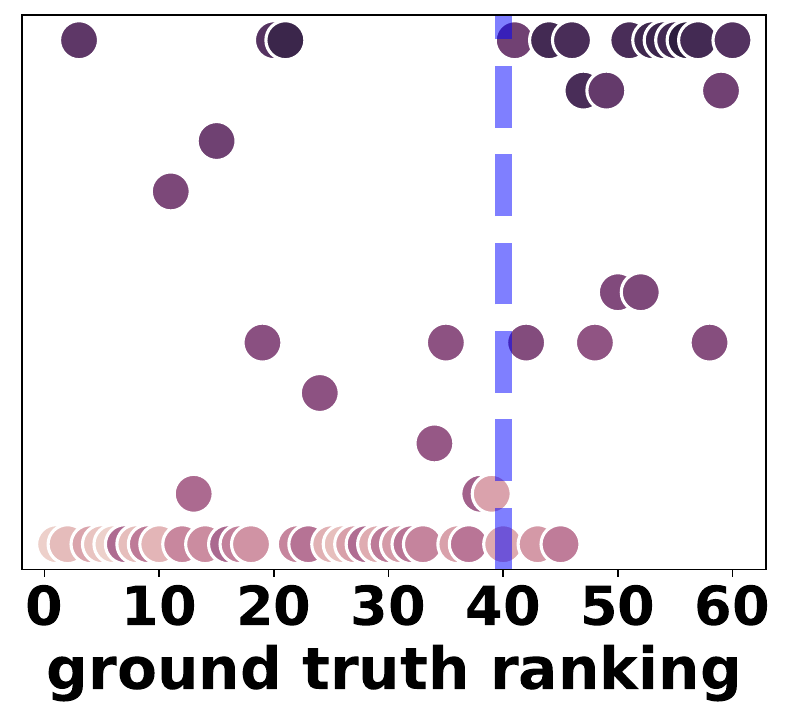}}
	\subfloat{\includegraphics[width = 0.14\linewidth, height=0.15\linewidth]{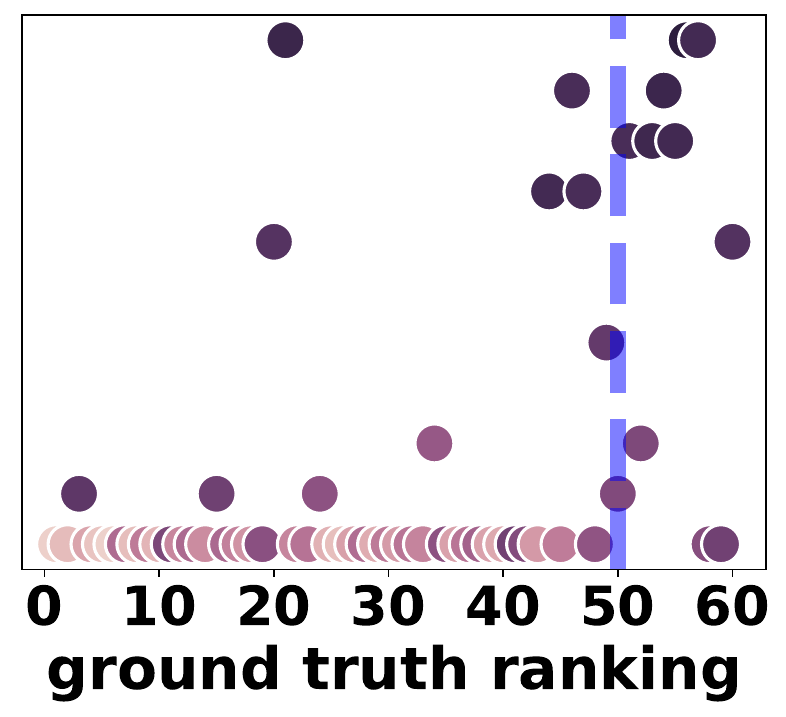}}
	\subfloat{\includegraphics[width = 0.14\linewidth, height=0.15\linewidth]{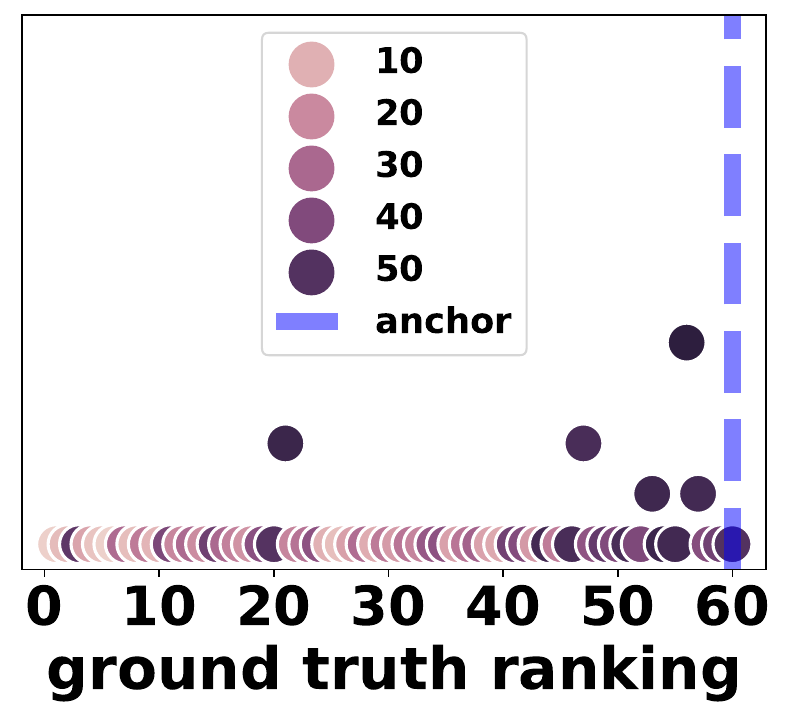}}

	\caption{Validating ranking uncertainty: The plots correspond to anchor images $x_a$ the $1^{st}, 10^{th}, 20^{th}, 30^{th}, 40^{th}, 50^{th}$, shown in blue -- with queries $x_q$ of systematically increasing rank (dots -- color indicates predicted rank, $y$-axis indicates confidence).}
 \label{fig:ranking_uncertainty}
\end{figure*}

\subsection{Validating Ranking Performance}
\paragraph*{Baselines} Six existing methods were used as baselines: RankNet~\cite{burges2005learning}, Hinge loss~\cite{yang2016deep}, ListNet~\cite{cao2007learning}, SoDeep~\cite{engilberge2019sodeep}, along with the ‘traditional’ image quality assessment method RankIQA~\cite{liu2017rankiqa} and a standard regression model using an L1 loss to regress the percentile rank of the image. We constructed the baseline model of a three-layer MLP including 512, 128, and 1 neurons in each hidden layer, which was trained in mini-batches. RankNet and Hinge loss were applied to all the image pairs within a batch. ListNet optimized the permutation probability distribution of the predicted scores to ground truth rankings. We trained two ListNet models: 'ListNet-local' was optimized to the rankings of samples within the minibatch, and 'ListNet-global' optimized to the global rankings of samples within the full training set. We trained an LSTM as the sorter for SoDeep. All the baseline models were trained with their optimal hyperparameter settings.

\begin{table}
    \caption{Performance on the clean test set. Here, NDCG@n means top-n normalized discounted cumulative gain.}
    \label{tab:clean}
    \centering
    \resizebox{\columnwidth}{!}{\begin{tabular}{lcccccc}
        \toprule
        Method     & SPC & PAcc & PRC & KTC & NDCG@3 & NDCG@5\\
        \midrule
        RankNet & $0.594$ & $71.9$ & $0.616$ & $0.437$ & $\textbf{0.881}$ & $\textbf{0.952}$\\
        Hinge loss & $0.602$ & $71.7$ & $0.604$ & $0.434$ & $0.790$ & $0.936$\\
        ListNet-loc. & $0.588$ & $70.2$ & $0.602$ & $0.405$ & $0.675$ & $0.740$ \\
        ListNet-glob. & $0.585$ & $70.1$ & $0.599$ & $0.402$ & $0.676$ & $0.742$\\
        SoDeep & $0.613$ & $71.9$ & $0.639$ & $0.438$ & $0.746$ & $0.841$\\
        RankIQA & $0.065$ & $50.2$ & $0.124$ & $0.003$ & $0.538$ & $0.649$\\
        Regression & $0.210$ & $57.1$ & $0.195$ & $0.141$ & $0.271$ & $0.287$\\
        \midrule
        ORBNet (ours) & $\textbf{0.652}$ & $\textbf{73.8}$ & $\textbf{0.673}$ & $\textbf{0.477}$ & $0.837$ & $0.925$ \\
        \bottomrule
    \end{tabular}}
\end{table}

\paragraph*{Validation on a cleaned test set} To minimize the effect of noise on validation, we first validate model performance on 60 head images annotated with scores by annotator A1 repeated at two different time points, which were averaged to produce a robust annotator rank. Tab.~\ref{tab:clean} shows performance evaluated with five metrics: Spearman correlation of ground truth and predicted rankings (SPC), overall classification accuracy of every pair in the dataset (PAcc), Pearson correlation of predicted and annotated image scores (PRC), Kendall's $\tau$ coefficient (KTC), and the top-3 and top-5 normalized discounted cumulative gain. ORBNet outperforms all baselines on SPC, PAcc, PRC, and KTC. On NDCG, it is slightly outperformed by RankNet and Hinge Loss but outperforms the remaining baselines by far.

\paragraph*{10-fold cross-validation} We also conducted a 10-fold cross-validation on all the anatomies, assessing model performance via Spearman correlation, see Tab.~\ref{tab:cv}. ORBNet outperforms baselines considerably on Head and Abdomen.

\begin{table}
  \caption{Spearman correlation of ground truth and predicted ranking under 10-fold cross-validation. }
  \label{tab:cv}
  \centering
  \begin{tabular}{lccc}
    \toprule
    Method     & Head     & Abdomen & Femur \\
    \midrule
    RankNet & $0.68 \pm 0.07$ & $0.58 \pm 0.12$ & $0.42 \pm 0.13$ \\
    Hinge loss & $0.70 \pm 0.06$ & $0.58 \pm 0.13$ & $0.42 \pm 0.15$\\
    ListNet-local & $0.54 \pm 0.12$ & $0.50 \pm 0.14$ & $\textbf{0.50}\pm 0.14$\\
    ListNet-global & $0.53 \pm 0.14$ & $0.51 \pm 0.13$ & $0.32 \pm 0.16$\\
    SoDeep & $0.66\pm 0.13 $& $0.58 \pm 0.13$ & $0.45\pm 0.15$\\
    RankIQA & $0.01 \pm 0.07$ & $-0.22 \pm 0.11$ & $0.20 \pm 0.06$\\
    Regression & $-0.02 \pm 0.18$ & $0.09 \pm 0.19$ & $0.07 \pm 0.15$\\
    \midrule
    ORBNet (ours) & $\textbf{0.81} \pm 0.09$ & $\textbf{0.67} \pm 0.15$ & $0.44 \pm 0.14$\\
    \bottomrule
  \end{tabular}
\end{table}

\paragraph*{Validating ranking uncertainty}

We compare a fixed anchor image $\mathbf{x}_a$ to query images $\mathbf{x}_q$ with ground truth ranking scores $y_q$ from the entire test set. Confidence was estimated using dropout probability 0.5 with 10 passes per query image. Fig.~\ref{fig:ranking_uncertainty} illustrates pairwise confidences of ranking $\mathbf{x}_a$ (blue dotted line) over a systematically varying $\mathbf{x}_q$ (dots), ordered by increasing query image rank, to show how ranking uncertainty depends on how similarly the anchor and query images are ranked by the annotator. This experiment is presented using the $1^{st}, 10^{th}, 20^{th}, 30^{th}, 40^{th}, 50^{th}$, and last ranked image in our clean test set as anchor image $x_a$. Note how, when the anchor image (dotted blue line) is ranked very high or low, most pairs have a high confidence ranking, whereas the confidence is lower -- and the uncertainty higher -- for neighboring images when the query image is ranked in the middle of the dataset.

\paragraph*{Model variance} To assess stability, we performed 10-fold cross-validation with 10 different random seeds. Tab.~\ref{tab:variance} shows the mean and standard deviation of SPC across the $100$ models per anatomy, indicating relatively robust ranking performance.

\begin{table}
  \caption{Mean and standard deviation of SPC of the ground truth and predicted image ranking in 10-fold cross validation.}
  \label{tab:variance}
  \centering
  \begin{tabular}{ccc}
    \toprule
    Head & Abdomen & Femur \\
    \midrule
    $0.7987\pm 0.09$ & $0.6648\pm 0.16$ & $0.4369\pm 0.14$ \\
    \bottomrule
  \end{tabular}
\end{table}

\section{Discussion, Limitations, and Conclusion}
\label{sec:discussion}

Our model is built on the assumption that annotated image quality rankings are noisy. This is confirmed by the intra-rater consistency for annotator A1 which gives an SPC of $0.794$ between the two repeated rankings of the $n=60$ head image dataset and an inter-rater consistency of $0.549$ with annotator A2. Thus, while annotations were performed by experts and based on the same criteria, both intra- and inter-rater consistency is limited, emphasizing that fine-grained semantic image quality assessment is inherently noisy and subjective. 

\paragraph*{Efficient ranking annotation via merge sort} Our proposed annotation scheme effectively achieves a complete ranking of the dataset. It does not, however, alleviate annotation noise. In part, images that are difficult to compare to each other should ideally be oversampled to reduce annotation noise.
Further, as we only compare image pairs once, we propagate annotation errors as later rankings depend on wrong decisions. Future work should therefore consider noise reduction while keeping the annotation process efficient for straightforward comparisons.

\paragraph*{Annotation uncertainty}
While we quantify ranking uncertainty in this study, we have not explored or modeled the underlying sources of uncertainty. In particular, our consistency assessment indicates aleatoric uncertainty stemming from the annotations, while MC dropout only captures epistemic uncertainty. Thus, discovering and explicitly modeling different sources of uncertainty are important open problems.

\paragraph*{Conclusion}
We address semantic image quality assessment via image ranking for a challenging fetal ultrasound task. Fetal ultrasound images are usually assessed as either acceptable or not. This goal may be unattainable in clinical practice, but we still want the best image quality possible. Thus, we have introduced a simple scheme for annotating global ranking of image sets. We propose an ordinal-regression-based neural network to learn image ranking in a coarse-to-fine manner, and utilize MC Dropout to quantify ranking uncertainty. We show that our model outperforms baselines in the challenging fetal ultrasound quality assessment task.  

\section*{Compliance with ethical standards}
This study is part of a larger project on AI support for clinicians carrying out fetal ultrasound examinations. Approvals from The Danish Data Protection Agency (Protocol No. P-2019-310) and by The Danish Patient Safety Authority (Protocol No. 3-3031-2915/1) have been obtained for the project. The project was submitted to the Regional Ethics Committee, which has assessed that the study is exempt from The Scientific Ethical Treatment Act (jr. nr. 21024445).

\section*{Acknowledgment}
This work was supported by the DIREC project \mbox{EXPLAIN-ME} (9142-00001B), the Novo Nordisk Foundation through the Center for Basic Machine Learning Research in Life Science (NNF20OC0062606), and the \mbox{Pioneer Centre for AI}, DNRF grant nr P1.

\bibliographystyle{IEEEtran}
\bibliography{ref}

\end{document}